\documentclass[letterpaper]{article} % DO NOT CHANGE THIS
\usepackage[preprint]{aaai2027}  % DO NOT CHANGE THIS
\usepackage[hyphens]{url}  % DO NOT CHANGE THIS
\usepackage{graphicx} % DO NOT CHANGE THIS
\usepackage{natbib}  % DO NOT CHANGE THIS AND DO NOT ADD ANY OPTIONS TO IT
\usepackage{caption} % DO NOT CHANGE THIS AND DO NOT ADD ANY OPTIONS TO IT
\usepackage{algorithm}
\usepackage{algorithmic}

\usepackage{newfloat}
\usepackage{listings}
\DeclareCaptionStyle{ruled}{labelfont=normalfont,labelsep=colon,strut=off} % DO NOT CHANGE THIS
\floatstyle{ruled}
\newfloat{listing}{tb}{lst}{}
\floatname{listing}{Listing}

\usepackage{booktabs}

\usepackage{amsfonts}
\usepackage{amsmath}
\usepackage{bbold}
\usepackage{amsthm}

\title{Every Wrong Answer Counts: Option-Level Psychometrics\\for LLM Multiple-Choice Benchmarks}
\author{
    Xiao Fei\textsuperscript{\rm 1}, Yang Zhang\textsuperscript{\rm 1}, Sarah Almeida Carneiro\textsuperscript{\rm 1}, Michalis Vazirgiannis\textsuperscript{\rm 1,\rm 2}\corresponding
}
\affiliations{
    \textsuperscript{\rm 1}École Polytechnique, Institut Polytechnique de Paris, France\\
    \textsuperscript{\rm 2}Mohamed bin Zayed University of Artificial Intelligence, United Arab Emirates\\
    xiao.fei@polytechnique.edu\\
    mvazirg@lix.polytechnique.fr
}

\begin{document}

\maketitle

\begin{abstract}

Most multiple-choice question (MCQ) benchmarks evaluate Large Language Models (LLMs) only by whether they select the correct answers. This binary scoring treats all incorrect responses alike, even though an LLM's preferences among incorrect options may contain systematic and useful information about its behavior and ability. We introduce the LLM Nominal Response Model (LLM-NRM), an option-aware psychometric framework that models the full distribution over answer choices to jointly estimate LLM ability and option-level item characteristics, while separating model-specific response calibration sharpness, positional preference, and difficulty-dependent fallback behavior. Across 189 LLMs and 31,554 items from 14 benchmarks, LLM-NRM predicts held-out LLM-item interactions more accurately than binary Item Response models and conventional nominal-response baselines, and its ability estimates achieve the strongest Spearman correlation of 0.920 with the external human-preference Arena.ai Elo leaderboard. Distractor identity contributes +101\% additional Fisher Information per item beyond correctness, and incorrect responses alone recover full-information ability estimates with Spearman 0.943. The learned item parameters also enable efficient benchmarking, where 41 selected items preserve the full-bank ranking with Kendall's correlation 0.85, corresponding to a 770 times reduction. In conclusion, we show that incorrect answers carry distinct and useful measurement information rather than representing equivalent mistakes.

% \color{green}{
% LLM -> better evaluation of LLM ability

% 1. Modeling likelihood is higher

% 2. Positional bias modeling and quantitative analysis

% 3. Prediction accuracy is higher

% 4. Correlation with Arena Elo is higher

% BENCHMARK -> better evaluation of benchmark quality

% 1. Benchmark item-side results (distractor modeling)

% 2. Super efficient sub-benchmark (Arena correlation, cost comparison, accuracy on 100+ items)

% 3. Benchmark quality quantitative results and comparison
% }
\end{abstract}


\section{Introduction}

\begin{figure}[t!]
    \centering
    \includegraphics[width=0.95\columnwidth]{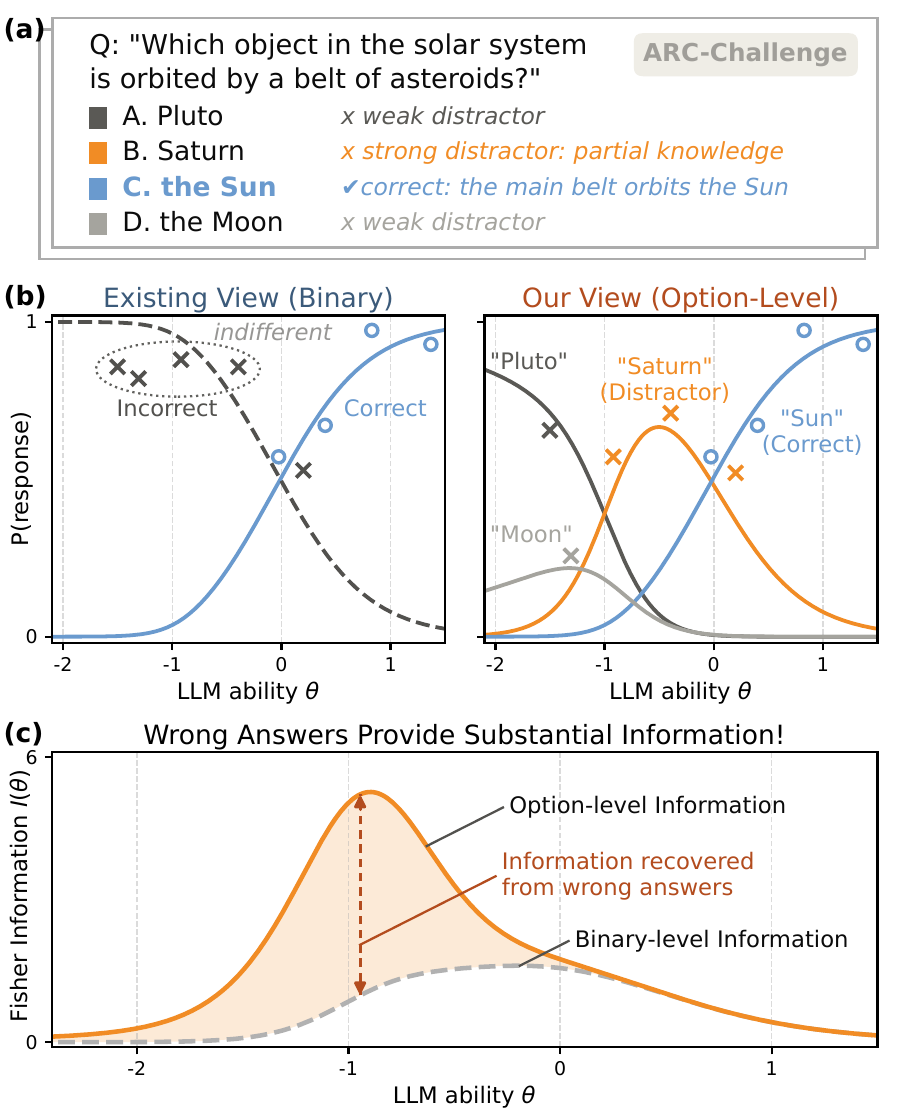}
    \caption{\textbf{A real example} of an ARC-Challenge MCQ item with item characteristic curves from our LLM-NRM fitting. (a) In this item, "Saturn" is a plausible distractor that reflects partial knowledge of LLM examinees, whereas "the Sun" is the correct answer. (b) Binary scoring collapses all three choices into the same "incorrect" outcome, but option-level modeling estimates a separate response curve instead for each option across LLM ability spectrum $\theta$. (c) Because these distractor preferences vary with ability, their identities provide Fisher information that correctness does not retain. }
    \label{fig:intro}
\end{figure}

Large Language Model (LLM) benchmarking commonly evaluates performance on standard multiple-choice question (MCQ) benchmarks with accuracy scores~\citep{wang2024mmlupro,rein2024gpqa}. However, accuracy is a limited measure of model capability, because it is tied to a particular benchmark: raw scores depend on its item composition, while empirical item difficulty depends on the evaluated model population. As a result, model rankings can shift with changes in the benchmark, scoring protocol, or comparison set~\citep{lalor2016building,perlitz2024efficient,alzahrani2024benchmarks}. A single accuracy value also collapses every response to a binary outcome, hiding model confidence and systematic error patterns~\citep{zhu2023calibration}. It does not by itself quantify measurement uncertainty~\citep{perlitz2024efficient}, and it loses discriminative power as the strongest models approach benchmark saturation~\citep{wang2024mmlupro}.

Item Response Theory (IRT) provides a principled framework for addressing some of these limitations by modeling interactions between latent ability and item characteristics~\citep{rasch1960,lord1968,birnbaum1968}. Recent studies apply psychometric models to LLM evaluation for benchmark analysis, adaptive testing, difficulty estimation, and multilingual evaluation~\citep{zhou2026lost,land2026auditing,zhuang2023efficiently,li2025adaptive,zhu2025llm,lior2026extending,zhang2026masked}. These approaches reveal benchmark saturation, item quality issues, and differences between models with similar accuracy scores~\citep{vania2021comparing,zhou2026lost,land2026auditing}. However, they generally reduce responses to binary correctness alone, discarding the option-level information contained in multiple-choice predictions.

In the meantime, LLM responses contain richer behavioral signals beyond correctness, including option-selection biases, positional preferences, self-evaluation, and response consistency patterns~\citep{zheng2024large,yang2025option,kadavath2022language,wu2025estimating,chaudhury2026quantifying}. Yet, these signals are typically analyzed separately from latent ability and item characteristics, leaving the full response structure underutilized.

In this paper, we demonstrate that the probability distribution over multiple-choice options produced by an LLM is itself a psychometric response. This naturally connects to the human-testing Nominal Response Model (NRM), a categorical extension of IRT that models relationships between latent ability and response option preferences~\citep{bock1972}. Unlike binary IRT, which scores each response only by correctness, the NRM gives every option its own discrimination and attractiveness, capturing how each distractor's appeal varies with ability and thereby extracting more measurement information per item than correctness alone.

We propose the LLM Nominal Response Model (LLM-NRM), which adapts NRM for LLM evaluation by modeling latent ability, item characteristics, and LLM-specific behavioral factors. Our contributions are as follows:

\begin{itemize}

\item To our knowledge, this work presents the first application of the Nominal Response Model (NRM) to LLMs, revealing latent response preferences across benchmark MCQ answer options and providing a richer view of model behavior beyond answer correctness;

\item We propose an LLM-adapted NRM formulation that incorporates behavioral factors such as calibration sharpness, ability-gated guessing, and positional bias, moving beyond traditional item-response modeling toward a richer representation of LLM decision processes; 

\item Through large-scale experiments on 189 LLMs and 31,554 MCQ items across 14 benchmarks, we show that LLM-NRM improves held-out response prediction, yields ability estimates that best match external human-preference rankings, and establishes that distractor choices are informative enough to recover ability and compress benchmarks by orders of magnitude.

\end{itemize}

\section{LLM Nominal Response Model}

We introduce the LLM Nominal Response Model (LLM-NRM), an option-aware extension of Bock's NRM for analyzing LLM responses to multiple-choice questions. Unlike conventional evaluation, which retains only the correctness of the most probable option, LLM-NRM models the complete response distribution over all valid options. This preserves richer information about distractor preferences and how option attractiveness varies across models of different abilities.

LLM-NRM retains the original NRM parameterization for MCQ items~\citep{penfield2014} and instead augments the respondent model with additional latent response processes motivated by empirical characteristics commonly observed in LLMs: (i) response calibration sharpness, (ii) positional preference, and (iii) difficulty-dependent fallback behavior. These LLM-specific extensions separately capture systematic response variation that would otherwise be absorbed into examinee ability and item modeling. Unlike human testing that typically records one categorical response per item~\citep{thissen1984,suh2010}, LLM evaluation provides full option distributions across large item banks, enabling these respondent-specific processes to be estimated.

\subsection{Problem Formulation}

Let $\mathcal{J}$ denote a collection of LLMs as examinees and $\mathcal{I}$ denote a collection of MCQ items. Each MCQ item $i\in\mathcal{I}$ presents $K_i$ answer options indexed by $\mathcal{K}_i=\{1,\ldots,K_i\}$, corresponding to the displayed answer positions such as A, B, C and D in the prompt. Let $g_i\in\mathcal{K}_i$ denote the index of the keyed correct option, and define $K_{\max}=\max_{i\in\mathcal{I}} K_i$ as the largest number of options among all items.

An LLM's response to an MCQ item is inherently distributional, as each autoregressive step produces a probability over next tokens. So when an LLM $j\in\mathcal{J}$ answers an MCQ item $i\in\mathcal{I}$, it expresses a degree of preference for every available option. We represent this observed preference $\mathbf{p}^{\mathrm{obs}}_{ji}$ as:
\begin{equation}
\mathbf{p}^{\mathrm{obs}}_{ji}=\left(p^{\mathrm{obs}}_{ji1},\ldots,p^{\mathrm{obs}}_{jiK_i}\right)\in\Delta^{K_i-1},
\end{equation}
with $p^{\mathrm{obs}}_{jik}$ the normalized preference for option $k$ and $\Delta^{K_i-1}$ the probability simplex.

LLM-NRM treats $\mathbf{p}^{\mathrm{obs}}_{ji}$ as the observed response, rather than reducing it to a binary indicator of whether the most probable option equals $g_i$.

\subsection{Bock's NRM}

Bock's NRM~\citep{bock1972} models unordered response options by assigning each examinee $j\in\mathcal{J}$ a latent ability $\theta_j\in\mathbb{R}$, and each option $k\in\mathcal{K}_i$ a discrimination parameter $a_{ik}\in\mathbb{R}$ plus an intercept $c_{ik}\in\mathbb{R}$. The discrimination controls how the option's relative attractiveness varies with ability, while the intercept captures its baseline attractiveness. The resulting response distribution is
\begin{equation}
p^{\mathrm{NRM}}_{jik}=\frac{\exp\left(a_{ik}\theta_j+c_{ik}\right)}{\sum_{k'\in\mathcal{K}_i}\exp\left(a_{ik'}\theta_j+c_{ik'}\right)}.
\end{equation}

Because the softmax is invariant to item-wise shifts in its utilities, the identification constraints are imposed
\begin{equation}
\sum_{k\in\mathcal{K}_i}a_{ik}=0, \qquad \sum_{k\in\mathcal{K}_i}c_{ik}=0,
\end{equation}
and anchor the ability scale with the prior $\theta_j\sim\mathcal{N}(0,1)$.

The classical NRM is powerful for human testing, but LLMs violate its assumptions in three systematic ways: 
\begin{itemize}
    \item differences in distribution sharpness that may reflect calibration rather than knowledge~\citep{kadavath2022language,zhu2023calibration},
    \item content-independent preferences for displayed positions~\citep{zheng2024large,pezeshkpour2024large,attali2003}, 
    \item and fallback guessing patterns on items that are difficult relative to the model's ability~\citep{ivgi2024fallback}.
\end{itemize}
Fitting NRM directly would absorb these non-ability effects into $\theta_j$ and the option parameters, potentially biasing both.

\subsection{Response Sharpness and Positional Bias}

We first augment the NRM ability-driven utilities with global response sharpness and content-independent positional preference.

\paragraph{Response Sharpness.}

Two LLMs with identical preference orderings over options may nevertheless produce distributions with substantially different concentrations because of differences in calibration rather than knowledge~\citep{kadavath2022language,zhu2023calibration}. Classical NRM has no examinee-specific parameter that changes concentration while preserving option ordering, so this variation may instead be absorbed into $\theta_j$ and the item discriminations.

We capture the global component of this variation with a positive sharpness parameter for each LLM: 
\begin{equation}
s_j\in\mathbb{R}^{+}. 
\end{equation}

Mathematically, $s_j$ is a per-LLM inverse temperature that changes distribution concentration without altering option ordering: $s_j>1$ sharpens the distribution, whereas $s_j<1$ flattens it. Thus, $s_j$ captures model-level calibration, or to what degree of confidence an LLM expresses its preferences, separately from item-specific knowledge.

\paragraph{Positional Bias.}

LLMs can exhibit systematic preferences for displayed answer positions or their labels independently of option content, and reordering the same options can substantially change measured accuracy~\citep{pezeshkpour2024large}. Because the NRM intercepts $c_{ik}$ are shared across LLMs, classical NRM cannot separate these model-specific positional effects from option attractiveness.

Rather than controlling for positional bias through computationally expensive permutation protocols~\citep{zheng2024large}, we estimate it jointly with ability: 
\begin{equation}
\boldsymbol{\delta}_j=\left(\delta_{j1},\ldots,\delta_{jK_{\max}}\right)\in\mathbb{R}^{K_{\max}},
\end{equation}
where $\delta_{jk}$ represents LLM $j$'s content-independent preference for displayed position $k$. 

Incorporating both model-specific sharpness and positional bias, we define the ability-driven response logit as
\begin{equation}
u_{jik}=s_j\left(a_{ik}\theta_j+c_{ik}+\delta_{jk}\right).
\end{equation}

\subsection{Difficulty-Gated Guessing Fallback}

Classical NRM represents all responses through the same ability-driven option utilities. When an item is difficult relative to an LLM's ability, however, its option preferences may increasingly reflect model-specific fallback strategies, such as label priors or option heuristics, rather than the item's content~\citep{ivgi2024fallback,zheng2024large,balepur2024}. Binary 3PL-IRT introduces a per-item lower asymptote~\citep{birnbaum1968}, but it neither models a complete fallback distribution over options nor allows fallback behavior to vary across LLMs, and its guessing parameters are usually poorly identified~\citep{barton1981,maris2009,sanmartin2015}.

We therefore define the guessing fallback gate instead: 
\begin{equation}
G_{ji}=\sigma\!\left[w_0-\kappa\left(\theta_j-\widetilde{b}_i\right)\right],
\end{equation}
where $w_0\in\mathbb{R}$ is a global intercept, $\kappa\geq 0$ is a global slope, and $\widetilde{b}_i$ is the item-difficulty index defined below. The logistic link $\sigma$ maps the ability-difficulty gap to a fallback interpolation weight in $(0,1)$. The intercept $w_0$ sets the transition location, while $\kappa$ controls its sharpness. Consequently, fallback becomes more prominent as item difficulty increases relative to LLM ability.

\subsubsection{An Approximation of Difficulty Index.}

The fallback gate requires a scalar measure of item difficulty $\tilde{b}_i$ derived from the NRM option parameters. In 2PL-IRT, item difficulty is defined as the ability at which the correct-response probability equals $\tfrac{1}{2}$. NRM has no explicit scalar difficulty parameter, but an analogous difficulty threshold can be expressed through the correct-versus-incorrect log-odds margin
\begin{equation}
f_i(\theta)=a_{ig_i}\theta+c_{ig_i}-\log\!\!\sum_{k\neq g_i}\!\exp\left(a_{ik}\theta+c_{ik}\right).
\end{equation}

Solving for this threshold numerically during optimization would be costly. We therefore construct a local and differentiable index by linearizing $f_i$ around $\theta=0$, the center of the ability prior:  
\begin{equation}
f_i(\theta)\approx f_i(0)+f_i'(0)\cdot \theta
\end{equation}
Thus, 
\begin{equation}
\widetilde{b}_i=-\frac{f_i(0)}{f_i'(0)}=\frac{\log\sum_{k\neq g_i}\exp(c_{ik})-c_{ig_i}}{a_{ig_i}-\sum_{k\neq g_i}w_{ik}a_{ik}},
\end{equation}
where
\begin{equation}
w_{ik}=\frac{\exp(c_{ik})}{\sum_{k'\neq g_i}\exp(c_{ik'})}.
\end{equation}
Here, $w_{ik}$ is the distractor softmax weight at $\theta=0$. Equivalently, $\widetilde{b}_i$ is the local threshold estimate obtained by taking one Newton step from the center of the ability prior. This construction introduces no additional free item parameter and remains differentiable during optimization.

\subsubsection{Fallback pattern.}

The scalar gate $G_{ji}$ determines how strongly fallback behavior contributes to a response. Inspired by respondent-specific bias vectors in human annotation models~\citep{dawid1979,welinder2010}, we assign each LLM a vector of fallback logits to specify the distribution over options within that regime 
\begin{equation}
\boldsymbol{\rho}_j=\left(\rho_{j1},\ldots,\rho_{jK_{\max}}\right)\in\mathbb{R}^{K_{\max}}, 
\end{equation}
where $\rho_{jk}$ is LLM $j$'s fallback logit for displayed position $k$. 

Notably, the positional-bias vector $\boldsymbol{\delta}_j$ and fallback vector $\boldsymbol{\rho}_j$ play distinct roles. The former captures a persistent positional effect in the ability-driven utilities, whereas the latter defines the response pattern that receives increasing weight as $G_{ji}$ grows.

\subsection{Full LLM-NRM Response Distribution}

Altogether, we define the response preference distribution under LLM-NRM as an interpolation combining the ability-driven and fallback processes:
\begin{equation}
\boxed{p^{\mathrm{LLM\text{-}NRM}}_{jik}=\frac{\exp\left[(1-G_{ji})\,u_{jik}+G_{ji}\,\rho_{jk}\right]}{\sum_{k'\in\mathcal{K}_i}\exp\left[(1-G_{ji})\,u_{jik'}+G_{ji}\,\rho_{jk'}\right]}}.
\end{equation}

Notably, with $s_j=1$, $\boldsymbol{\delta}_j=\boldsymbol{\rho}_j=\mathbf{0}$, and $G_{ji}=0$, our modeling reduces exactly to the classical NRM. 

Because the softmax is invariant to adding the same constant to all logits, we impose the LLM-level identification constraints: 
\begin{equation}
\sum_{k\leq K_{\max}}\delta_{jk}=0,\text{ and }\sum_{k\leq K_{\max}}\rho_{jk}=0, 
\end{equation}
and positivity are enforced through unconstrained reparameterizations:
$s_j=\operatorname{softplus}(\bar{s}_j)$, $\kappa=\operatorname{softplus}(\bar{\kappa})$, with $\bar{s}_j,\bar{\kappa}\in\mathbb{R}$.

\subsection{Fitting}

In total, LLM-NRM estimates $2(K_i-1)$ free parameters per MCQ item, plus $2K_{\max}$ free scalars per LLM along with two more global scalars $(w_0,\kappa)$. Given a group of LLMs $\mathcal{J}$ and their observed preference $\mathbf{p}^{\mathrm{obs}}_{ji}$ on a set of MCQ items $\mathcal{I}$, we fit all parameters jointly by maximum a posteriori (MAP) estimation with Adam optimizer on the soft cross-entropy loss: 
\begin{equation}
\mathcal{L}=-\sum_{i\in \mathcal{I}, j\in \mathcal{J},k\in\mathcal{K}_i}p^{\mathrm{obs}}_{jik}\,\log p^{\mathrm{LLM\text{-}NRM}}_{jik}+\frac{1}{2}\sum_{j\in\mathcal{J}}\theta_j^{2},
\end{equation}
where $\frac{1}{2}\theta_j^{2}$ is the regularization term of the prior $\theta\sim\mathcal{N}(0,1)$.

\section{Experimental Setup}

\paragraph{MCQ item bank.}
We assemble a large and diverse evaluation set comprising \emph{31,554 MCQ items} from 14 benchmarks spanning factual knowledge, reasoning, and commonsense: MMLU-Pro~\citep{wang2024mmlupro}, GPQA-Diamond~\citep{rein2024gpqa}, ARC-Challenge~\citep{clark2018arc}, AGIEval~\citep{zhong2024agieval}, CommonsenseQA~\citep{talmor2019commonsenseqa}, LogiQA 2.0~\citep{liu2023logiqa2}, OpenBookQA~\citep{mihaylov2018openbookqa}, TruthfulQA~\citep{lin2022truthfulqa}, MedQA-USMLE~\citep{jin2021medqa}, RACE~\citep{lai2017race}, SocialIQA~\citep{sap2019social}, QASC~\citep{khot2020qasc}, ReClor~\citep{yu2020reclor}, and MMLU~\citep{hendrycks2021mmlu}. Because MMLU and MMLU-Pro partially overlap, we remove from MMLU every MCQ item that also appears in MMLU-Pro. The resulting collection contains between 2 and 10 answer options per MCQ item, with a mean of 5.82. This variation allows us to evaluate LLM-NRM across both conventional and large-option MCQ formats.

\paragraph{LLM fleet.}
Our examinee fleet contains \emph{189 LLMs}, accessed through either publicly released weights or official APIs. We intentionally cover a broad range of LLM ability, parameter scale, training recipe, architecture, and development period. The collection includes 64 open-weight LLM families, including Qwen, Llama, Gemma, Phi, Mistral, Falcon, and Pythia, with parameter counts ranging from 70 million to 235 billion and release dates from February 2019 to May 2026. Different model architectures are also covered, including dense Transformers, mixture-of-experts models, and state-space models. API-accessible systems including DeepSeek-V4 Pro, GLM-5.1, and Kimi-K2.6 extend the fleet to recent frontier-scale LLMs. 

\paragraph{Answer-choice distributions.}
For each LLM--MCQ item pair, we extract the log-probability assigned to every valid answer option at the first token position immediately following the question and answer instruction. All experiments apply common settings in LLM MCQ benchmarking, using zero-shot prompting, temperature $T=1$, with chain-of-thought disabled. We compute option scores from the full-vocabulary log-softmax whenever complete logits are available. Some APIs expose only the top-20 next-token logits where we retain valid option tokens present in the returned set, and assign zero observed mass to unreported options. The global answer-instruction prompt is designed to concentrate next-token mass on option tokens, with only 0.56\% letter-mass leakage reported across the entire fleet.

\paragraph{Baselines.}
We compare LLM-NRM with classical binary 1PL--4PL IRT~\citep{barton1981}, Bock's NRM~\citep{bock1972}, as well as prior state-of-the-art approaches including Deep-IRT~\citep{yeung2019deepirt}, $\beta^3$-IRT~\citep{chen2019beta}, PSN-IRT~\citep{zhou2026lost}, and SD-IR~\citep{metaeval2026}. Binary models natively predict only correctness, and whenever they are evaluated on complete responses, their predicted incorrect mass is distributed using an MCQ-item-specific wrong-option distribution estimated exclusively from training cells. All models are trained with Adam optimizer with learning rate 0.05 for 2,500 steps, under which a stable convergence is always reached.

\section{A Better Modeling of LLM Responses}

We organize our evaluation around three questions: (i) whether LLM-NRM better models LLM responses, (ii) how much information incorrect-option identity contributes beyond correctness, and (iii) whether this additional information enables more efficient benchmarking.

\paragraph{Held-out Response Prediction.}

\begin{table*}[t]
\centering
\begin{tabular}{@{}lcccccc@{}}
\toprule
 & & \multicolumn{3}{c}{binary correctness channel} & \multicolumn{2}{c}{answer option channel} \\
\cmidrule(lr){3-5}\cmidrule(l){6-7}
model & type & Acc $\uparrow$ & LL $\downarrow$ & Brier $\downarrow$ & Acc $\uparrow$ & LL $\downarrow$ \\
\midrule
1PL-IRT~\citep{rasch1960} & binary & 0.8091 {\scriptsize$\pm$.0001} & 0.4437 {\scriptsize$\pm$.0003} & 0.1404 {\scriptsize$\pm$.0001} & 0.6121 {\scriptsize$\pm$.0001} & 1.0480 {\scriptsize$\pm$.0012} \\
2PL-IRT~\citep{lord1968} & binary & 0.8163 {\scriptsize$\pm$.0002} & 0.3908 {\scriptsize$\pm$.0002} & 0.1256 {\scriptsize$\pm$.0001} & 0.6373 {\scriptsize$\pm$.0002} & 0.9949 {\scriptsize$\pm$.0012} \\
3PL-IRT~\citep{birnbaum1968} & binary & 0.8227 {\scriptsize$\pm$.0003} & 0.3839 {\scriptsize$\pm$.0003} & 0.1223 {\scriptsize$\pm$.0001} & 0.6417 {\scriptsize$\pm$.0002} & 0.9879 {\scriptsize$\pm$.0012} \\
4PL-IRT~\citep{barton1981} & binary & 0.8232 {\scriptsize$\pm$.0003} & 0.3847 {\scriptsize$\pm$.0005} & 0.1221 {\scriptsize$\pm$.0001} & 0.6427 {\scriptsize$\pm$.0003} & 0.9889 {\scriptsize$\pm$.0015} \\
Deep-IRT~\citep{yeung2019deepirt} & binary & 0.7980 {\scriptsize$\pm$.0003} & 0.4846 {\scriptsize$\pm$.0003} & 0.1555 {\scriptsize$\pm$.0001} & 0.5923 {\scriptsize$\pm$.0003} & 1.0890 {\scriptsize$\pm$.0011} \\
$\beta^3$-IRT~\citep{chen2019beta} & binary & 0.8049 {\scriptsize$\pm$.0004} & 0.4154 {\scriptsize$\pm$.0003} & 0.1351 {\scriptsize$\pm$.0001} & 0.6184 {\scriptsize$\pm$.0001} & 1.0199 {\scriptsize$\pm$.0009} \\
PSN-IRT~\citep{zhou2026lost} & binary & 0.8207 {\scriptsize$\pm$.0008} & 0.3877 {\scriptsize$\pm$.0014} & 0.1239 {\scriptsize$\pm$.0002} & 0.6388 {\scriptsize$\pm$.0004} & 0.9921 {\scriptsize$\pm$.0025} \\
SD-IR~\citep{metaeval2026} & option & 0.7077 {\scriptsize$\pm$.0001} & 0.6782 {\scriptsize$\pm$.0004} & 0.2430 {\scriptsize$\pm$.0002} & 0.5887 {\scriptsize$\pm$.0003} & 1.3774 {\scriptsize$\pm$.0003} \\
\midrule
NRM, soft channel & option & 0.8161 {\scriptsize$\pm$.0003} & 0.3893 {\scriptsize$\pm$.0001} & 0.1258 {\scriptsize$\pm$.0001} & 0.6567 {\scriptsize$\pm$.0003} & 0.9393 {\scriptsize$\pm$.0007} \\
NRM, hard channel~\citep{bock1972} & option & 0.8149 {\scriptsize$\pm$.0003} & 0.3931 {\scriptsize$\pm$.0001} & 0.1264 {\scriptsize$\pm$.0001} & 0.6698 {\scriptsize$\pm$.0001} & 0.9171 {\scriptsize$\pm$.0013} \\
\midrule
\textbf{LLM-NRM (full)} & option & \textbf{0.8395} {\scriptsize$\pm$.0002} & \underline{0.3542} {\scriptsize$\pm$.0004} & \textbf{0.1127} {\scriptsize$\pm$.0001} & \textbf{0.7070} {\scriptsize$\pm$.0003} & \textbf{0.8181} {\scriptsize$\pm$.0011} \\
\qquad w/o guessing fallback ($G,\boldsymbol{\rho}$) &  & 0.8380 {\scriptsize$\pm$.0002} & 0.3570 {\scriptsize$\pm$.0003} & 0.1137 {\scriptsize$\pm$.0001} & 0.7054 {\scriptsize$\pm$.0001} & 0.8198 {\scriptsize$\pm$.0006} \\
\qquad w/o calibration sharpness ($s$) &  & \underline{0.8383} {\scriptsize$\pm$.0002} & \textbf{0.3538} {\scriptsize$\pm$.0004} & \underline{0.1128} {\scriptsize$\pm$.0002} & 0.7042 {\scriptsize$\pm$.0003} & 0.8202 {\scriptsize$\pm$.0010} \\
\qquad w/o positional bias ($\boldsymbol{\delta}$) &  & 0.8382 {\scriptsize$\pm$.0003} & 0.3568 {\scriptsize$\pm$.0003} & 0.1136 {\scriptsize$\pm$.0001} & \underline{0.7056} {\scriptsize$\pm$.0003} & \underline{0.8194} {\scriptsize$\pm$.0007} \\
% \qquad $\boldsymbol{\rho}$ shared across LLMs &  & \underline{0.8389} {\scriptsize$\pm$.0001} & 0.3553 {\scriptsize$\pm$.0002} & 0.1130 {\scriptsize$\pm$.0001} & 0.7039 {\scriptsize$\pm$.0004} & 0.8228 {\scriptsize$\pm$.0008} \\
\bottomrule
\end{tabular}
\caption{Held-out response prediction under 5-fold cross-validation. Values report the mean and standard deviation across five disjoint test folds. We report top-1 accuracy (Acc) and log-loss (LL) for both response channels, together with the Brier score (Brier) for binary correctness. Arrows indicate the preferred direction.}
\label{tab:race}
\end{table*}

We evaluate held-out prediction of the LLM--item response matrix using 5-fold cell-level cross-validation. Table~\ref{tab:race} reports performance in two channels: binary correctness, where option-aware models predict the probability of the keyed answer; and answer-option prediction, where binary models distribute non-key probability mass across distractors using training-set frequencies.

The binary channel favors binary IRT baselines, yet LLM-NRM achieves the highest correctness accuracy among competing models ($0.8395$ vs. $0.8232$). In the answer-option channel, which directly evaluates option-level modeling, LLM-NRM provides larger gains, achieving $0.7070$ accuracy and $0.8181$ log-loss compared with $0.6427$ and $0.9879$ for the strongest binary baselines, and $0.6698$ and $0.9171$ for hard-channel NRM. Ablation results further show that all proposed components contribute in a complementary way, as removing all of them reduces LLM-NRM (full) to NRM (soft channel) with significant performance gap.

\paragraph{External Validity of the Ability Scale.}

\begin{table}[h]
\centering
\begin{tabular}{@{}lc@{}}
\toprule
measurement & Spearman \\
\midrule
Raw Accuracy & 0.866 {\scriptsize$\pm$.024} \\
1PL-IRT-$\theta$ & 0.866 {\scriptsize$\pm$.024} \\
2PL-IRT-$\theta$ & 0.857 {\scriptsize$\pm$.024} \\
3PL-IRT-$\theta$ & 0.874 {\scriptsize$\pm$.022} \\
4PL-IRT-$\theta$ & \underline{0.885} {\scriptsize$\pm$.021} \\
PSN-IRT-$\theta$ & 0.877 {\scriptsize$\pm$.022} \\
Deep-IRT-$\theta$ & 0.865 {\scriptsize$\pm$.024} \\
$\beta^3$-IRT-$\theta$ & 0.814 {\scriptsize$\pm$.032} \\
SD-IR-$\theta$ & 0.829 {\scriptsize$\pm$.029} \\
NRM (soft channel)-$\theta$ & 0.766 {\scriptsize$\pm$.037} \\
NRM (hard channel)-$\theta$ & 0.831 {\scriptsize$\pm$.029} \\
\textbf{LLM-NRM-$\theta$} & \textbf{0.920} {\scriptsize$\pm$\textbf{.015}} \\
\bottomrule
\end{tabular}
\caption{Rank agreement between each measurement and Arena.ai text Elo scores. Values report Spearman correlation with bootstrap standard deviation. LLM-NRM achieves the highest observed Spearman correlation.}
\label{tab:arena_elo_ext_rho}
\end{table}

Predictive performance alone does not guarantee that estimated LLM abilities capture meaningful model differences. We therefore compare LLM-NRM ability rankings with independent Arena.ai Text Elo scores, derived from crowdsourced pairwise human preferences on open-ended prompts~\citep{chiang2024chatbot}. This evaluates whether the learned scale aligns with external human judgments rather than benchmark accuracy alone. We match 48 LLMs with a July 2026 leaderboard snapshot\footnote{Arena.ai Text leaderboard: \url{https://arena.ai/leaderboard/text}.} and compute Spearman rank correlation, with uncertainty estimated by bootstrap resampling of the matched LLMs.

As shown in Table~\ref{tab:arena_elo_ext_rho}, LLM-NRM achieves the strongest agreement with Arena Elo ($\rho=0.920$), outperforming raw accuracy ($0.866$) and the strongest competing latent ability estimate ($0.885$). Since Arena outcomes are not used during training, this result indicates that option-level response patterns capture model differences aligned with human preferences beyond aggregate MCQ correctness. The automatically estimated scale provides a reproducible and lower-cost proxy for Arena-style ranking.

\section{Information Beyond Binary Correctness}

\begin{figure*}[t]
    \centering
    \begin{minipage}[t]{0.49\textwidth}
        \centering
        \includegraphics[width=0.98\linewidth]{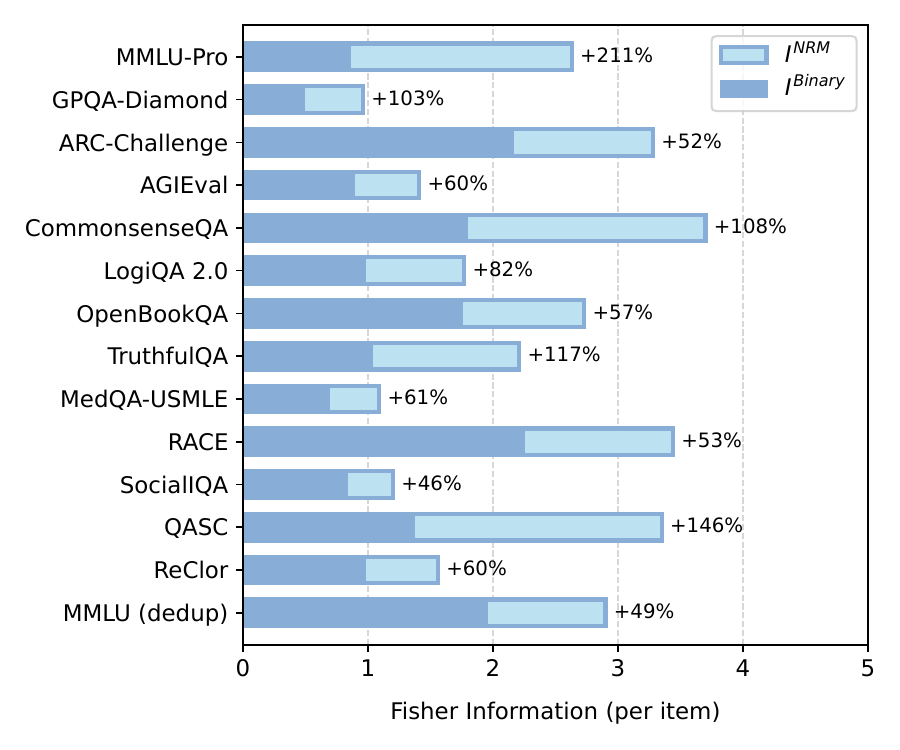}\\[-0.4em]
        \textbf{(a)} Information retained by the response channels
    \end{minipage}
    \hfill
    \begin{minipage}[t]{0.49\textwidth}
        \centering
        \includegraphics[width=0.98\linewidth]{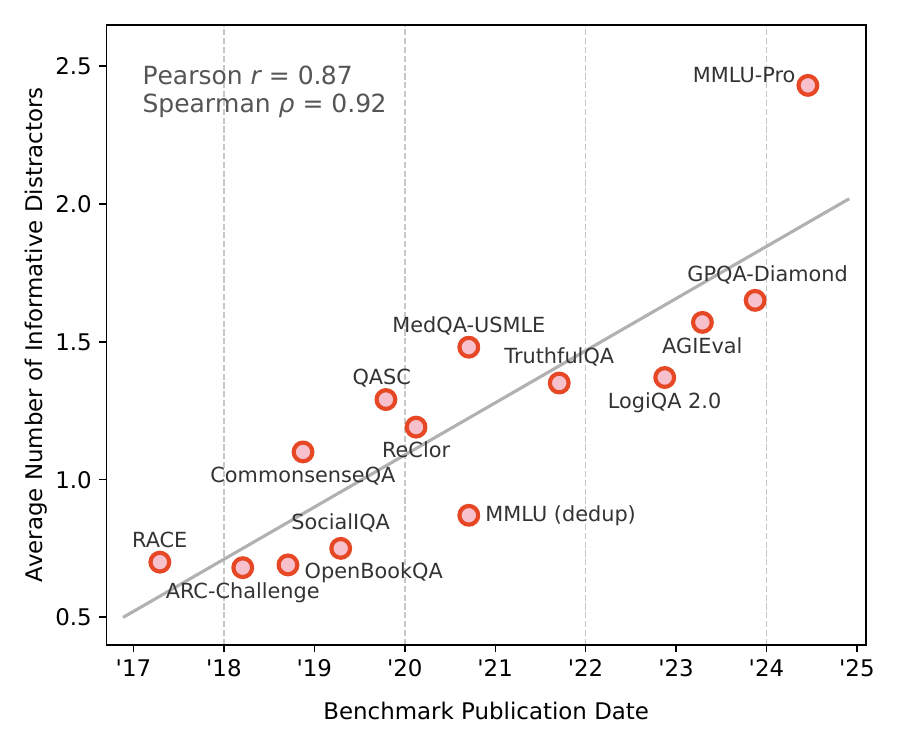}\\[-0.4em]
        \textbf{(b)} Informative distractors across benchmark dates
    \end{minipage}
    \caption{\textbf{Ability information beyond binary correctness.} \textbf{(a)} Mean item-level Fisher information in binary correctness and the additional information contributed by distractor identity. The full stacked length represents the information in the complete option response; percentages denote its increase over binary scoring. \textbf{(b)} Average number of detected informative distractors per item versus benchmark publication date. More recent benchmarks tend to expose more distinguishable distractor structure.}
    \label{fig:distractor-information}
\end{figure*}

We next examine the source of this predictive advantage of LLM-NRM: binary scoring observes only \emph{whether} an LLM is wrong, whereas option-level scoring also observes \emph{which} distractor it selects. We next demonstrate this additional information both theoretically and empirically. 

\paragraph{Fisher Information from Distractor Identity.}

For the nominal-response component, as commonly defined in prior works~\citep{bock1972,suh2010,garcia2014multiple}, the Fisher Information in the full option identity decomposes into the information retained by binary correctness and an additional nonnegative term contributed by distractor identity:

\begin{equation}
I^{\mathrm{NRM}}_i(\theta)=I^{\mathrm{binary}}_i(\theta)+I^{\mathrm{wrong}}_i(\theta), 
\end{equation}
where $I^{\mathrm{wrong}}_i(\theta) \geq 0.$

Figure~\ref{fig:distractor-information}(a) reports the item-level Fisher information retained by coarsened binary correctness and the additional contribution of distractor identity, averaged over the ability distribution and over items within each benchmark. The full option response provides $+101\%$ more Fisher information per item than binary scoring on average, and the gain is positive for every benchmark, showing that ability-dependent distractor preferences are a systematic property of LLM responses rather than an isolated feature of a few datasets.

\paragraph{Informativeness across benchmarks.} 

Figure~\ref{fig:distractor-information}(b) characterizes how the available distractor signal varies across benchmark designs, as we define an incorrect option as informative distractor if it's nevertheless the single most likely response at some ability level. More recently released benchmarks tend to contain more informative distractors per item, and many recent benchmarks have more than 1 informative distractor per item. It indicates that the response structure discarded by binary scoring remains substantial, and may become increasingly consequential as MCQ benchmarks incorporate richer sets of distractors.

\paragraph{Ability Estimation from Incorrect Responses.}

We next isolate the contribution of distractor identity with another 5-fold cross-validation over the LLM fleet. In each fold, MCQ item parameters are fit using the training LLMs and then held fixed while estimating $\theta$ for the held-out LLMs under two conditions: (i) Binary-Only, where the estimator observes only correctness signals; and (ii) Incorrect-Only, where it observes only the categorical signals from incorrectly answered MCQ items. For each held-out LLM, we compare the estimated $\theta$ with the reference estimate obtained using the full response information. Table~\ref{tab:wrong_only} reports the resulting correlations across all folds, quantifying the information contributed by each response signal.

\begin{table}[h]
\centering
\begin{tabular}{@{}lcc@{}}
\toprule
scenario & Pearson $\uparrow$ & Spearman $\uparrow$ \\
\midrule
Binary-Only & 0.8983 {\scriptsize$\pm$.0076} & 0.9785 {\scriptsize$\pm$.0075} \\
Incorrect-Only & 0.8693 {\scriptsize$\pm$.0615} & 0.9428 {\scriptsize$\pm$.0174} \\
\bottomrule
\end{tabular}
\caption{Correlation between held-out $\theta$ estimates under each observation scenario and the full-fleet reference estimate (mean {\scriptsize$\pm$sd} over 5-fold cross-validation).}
\label{tab:wrong_only}
\end{table}

The results confirm that distractor identity is a meaningful source of information for ability estimation. Even without observing whether an answer is correct, the pattern of selected distractors yields a strong estimate of $\theta$, indicating that error structure alone is highly informative but not a random effect. Neither scenario leads to a perfect correlation, suggesting that the two signals provide complementary information.

\section{Data Efficient LLM Benchmarking}

\begin{figure*}[t]
    \centering
    \begin{minipage}[t]{0.49\textwidth}
        \centering
        \includegraphics[width=0.98\linewidth]{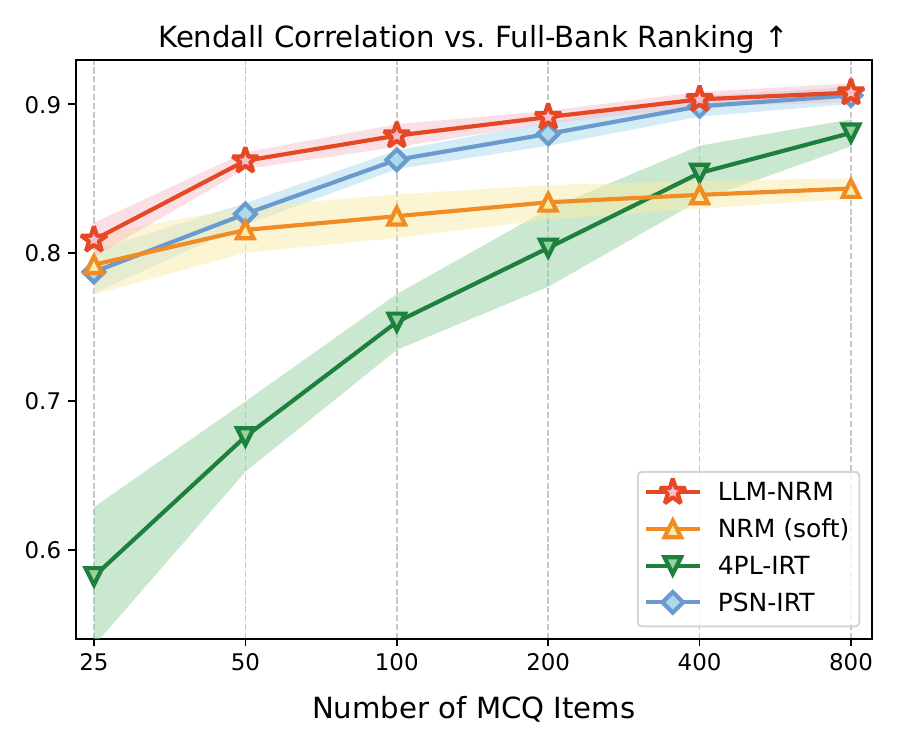}\\[-0.4em]
        \textbf{(a)} Ranking new LLMs with fewer MCQ items
    \end{minipage}
    \hfill
    \begin{minipage}[t]{0.49\textwidth}
        \centering
        \includegraphics[width=0.98\linewidth]{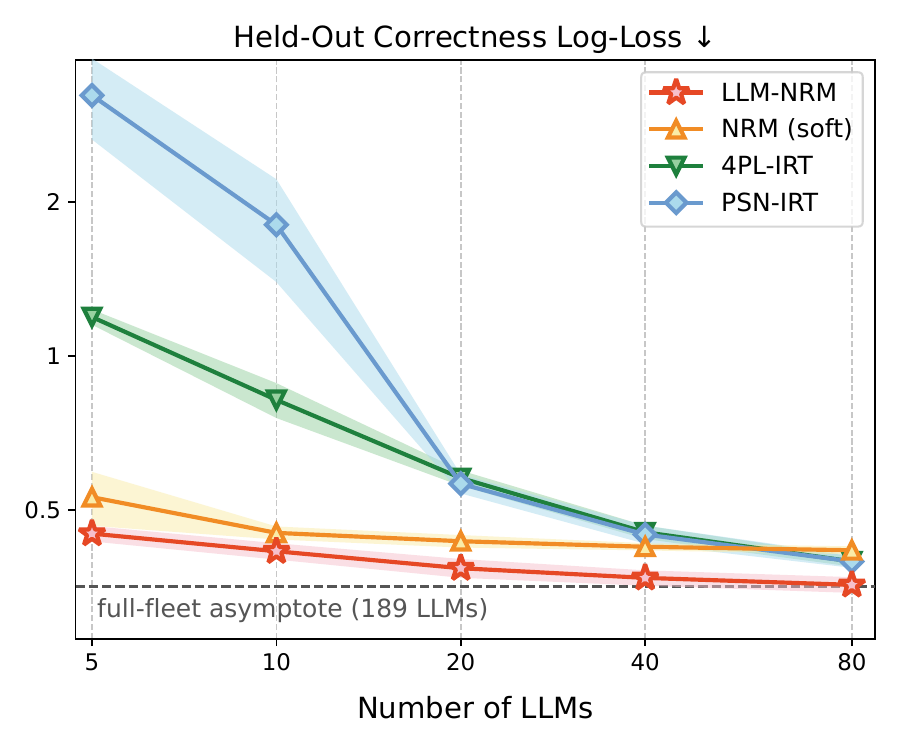}\\[-0.4em]
        \textbf{(b)} Calibrating an MCQ item bank with fewer LLMs
    \end{minipage}
    \caption{\textbf{Data-efficient LLM evaluation and benchmark calibration.} (a) Kendall correlation between rankings from information-selected subsets and the full item bank. LLM-NRM performs best, especially at small item budgets. (b) Held-out correctness log-loss after calibration with fewer LLMs. LLM-NRM produces useful item parameters with only 5 LLMs and approaches full-fleet performance near 40 MCQ items. Bands show variation across splits in (a) and calibration subsets in (b).}
    \label{fig:mpcurve}
\end{figure*}

As a natural consequence, we examine how LLM-NRM's richer response modeling enables more efficient LLM MCQ benchmarking. Prior compression methods are binary-correctness-based, typically need around 100 anchor items, and calibrate on thousands of models~\citep{polo2024tinybenchmarks,kipnis2025metabench}. We therefore investigate two questions as two typical costs of benchmarking on our approach: (i) how many MCQ items are required to reliably rank new LLMs? and (ii) how many LLMs are needed to calibrate a reusable MCQ item bank?

% As a natural consequence, we examine how LLM-NRM's more informative response modeling facilitates more efficient LLM MCQ benchmarking. The cost of LLM MCQ benchmarking has two complementary sources. Once an MCQ item bank has been calibrated, every new LLM must answer enough MCQ items to obtain a reliable LLM ability estimate. Before that bank can be used, enough calibration LLMs must answer its MCQ items to estimate stable MCQ item parameters. We therefore examine two problems: (i) How many MCQ items are needed to rank new LLMs?; (ii) How many LLMs are needed to calibrate a reusable MCQ item bank?

\paragraph{Compression of MCQ Benchmarks.}

We first vary the number of MCQ items administered to each held-out LLM. Each model uses its calibrated MCQ item parameters to select an information-rich subset of the specified size, then estimates LLM ability from only the selected responses. We evaluate the resulting leaderboard using Kendall's \(\tau\) with the ranking obtained from the complete MCQ item bank. 

Figure~\ref{fig:mpcurve}(a) shows that LLM-NRM remains effective even with a very small MCQ item budget. Using only the 41 most informative MCQ items, the reduced benchmark achieves a Kendall correlation of $0.85$, corresponding to a $\times770$ compression rate. Overall, LLM-NRM performs best in the low-budget regime, while PSN-IRT becomes competitive only when several hundred MCQ items are available. These results indicate that a new LLM can be faithfully calibrated by evaluating it on only a small set of MCQ items.

\paragraph{Calibration with Fewer LLMs.}

We next consider the reverse calibration problem. Unlike conventional psychometric IRT, which assumes a large population of examinees calibrating a relatively small item bank, LLM benchmarking often enters the opposite regime: tens of thousands of MCQ items but only tens or hundreds of LLMs for calibration. Flexible binary IRT models must therefore estimate item parameters from limited responses. We test whether LLM-NRM alleviates this bottleneck by randomly sampling calibration subsets, estimating item parameters, and evaluating the resulting MCQ banks on held-out LLMs with binary log loss on correctness for fair comparison.

Figure~\ref{fig:mpcurve}(b) shows that LLM-NRM consistently achieves the lowest held-out correctness loss across all calibration fleet sizes. Unlike binary IRT models, which become unstable with small calibration fleets due to limited information in binary responses, LLM-NRM exploits option-level response distributions to obtain richer and more data-efficient calibration signals. Compared with soft-channel NRM, LLM-NRM's modeling of LLM-specific response sharpness, positional bias, and guessing behavior further improves generalization to unseen LLMs. These results demonstrate that LLM-NRM reduces both the number of MCQ items required to evaluate new LLMs and the number of LLMs needed for benchmark calibration, lowering evaluation costs.

\section{Discussion and Conclusion}

We introduced LLM-NRM, an option-aware psychometric model that recovers information discarded by binary MCQ scoring. By modeling option distributions, LLM-NRM jointly estimates LLM ability, item characteristics, and behavioral factors such as response calibration sharpness, positional preference, and difficulty-dependent fallback behavior. Across 189 LLMs and 31,554 items from 14 benchmarks, it improves held-out response prediction, aligns ability estimates with human-preference rankings, and reveals that distractor choices provide substantial information beyond correctness. These signals also improve evaluation efficiency by preserving benchmark rankings with fewer items and reducing the number of LLMs required for calibration.

On the other hand, the current formulation assumes access to comparable option probabilities and models ability with a single latent dimension. Future work can explore partial probability observations, multidimensional abilities, and extensions beyond fixed-option questions. After all, our results demonstrate that LLM errors are informative signals that reveal their underlying capabilities and decision patterns, enabling more reliable evaluation beyond accuracy-based rankings.

\bibliography{aaai2027}

\end{document}